\documentclass{bmvc2k}

\title{Object Detection as a \\Positive-Unlabeled Problem}

\addauthor{Yuewei Yang*}{yuewei.yang@duke.edu}{1}
\addauthor{Kevin J Liang*}{kevin.liang@duke.edu}{1}
\addauthor{Lawrence Carin}{lcarin@duke.edu}{1}

\addinstitution{
 Duke University\\
 Durham, NC, USA \\
 \vspace{3mm}
 * Equal contribution
}

\runninghead{Yang, Liang, Carin}{Object Detection as a Positive-Unlabeled Problem}

\usepackage{amsmath}
\usepackage{amssymb}
\usepackage{array}
\usepackage{booktabs}
\usepackage{wrapfig}
\usepackage{float}
\newcolumntype{M}[1]{>{\centering\arraybackslash}m{#1}}
\DeclareMathOperator*{\argmax}{arg\!\max}

\begin{document}
\graphicspath{{images/}}

\maketitle

\begin{abstract}
	As with other deep learning methods, label quality is important for learning modern convolutional object detectors.
	However, the potentially large number and wide diversity of object instances that can be found in complex image scenes makes constituting complete annotations a challenging task. 
	Indeed, objects missing annotations can be observed in a variety of popular object detection datasets.
	These missing annotations can be problematic, as the standard cross-entropy loss employed to train object detection models treats classification as a positive-negative (PN) problem: unlabeled regions are implicitly assumed to be background.
	As such, any object missing a bounding box results in a confusing learning signal, the effects of which we observe empirically.
	To remedy this, we propose treating object detection as a positive-unlabeled (PU) problem, which removes the assumption that unlabeled regions must be negative.
	We demonstrate that our proposed PU classification loss outperforms the standard PN loss on PASCAL VOC and MS COCO across a range of label missingness, as well as on Visual Genome and DeepLesion with full labels.   
\end{abstract}

\section{Introduction}
\label{sec:intro}
\begin{figure}[t]
	\centering
	\includegraphics[width=0.25\textwidth]{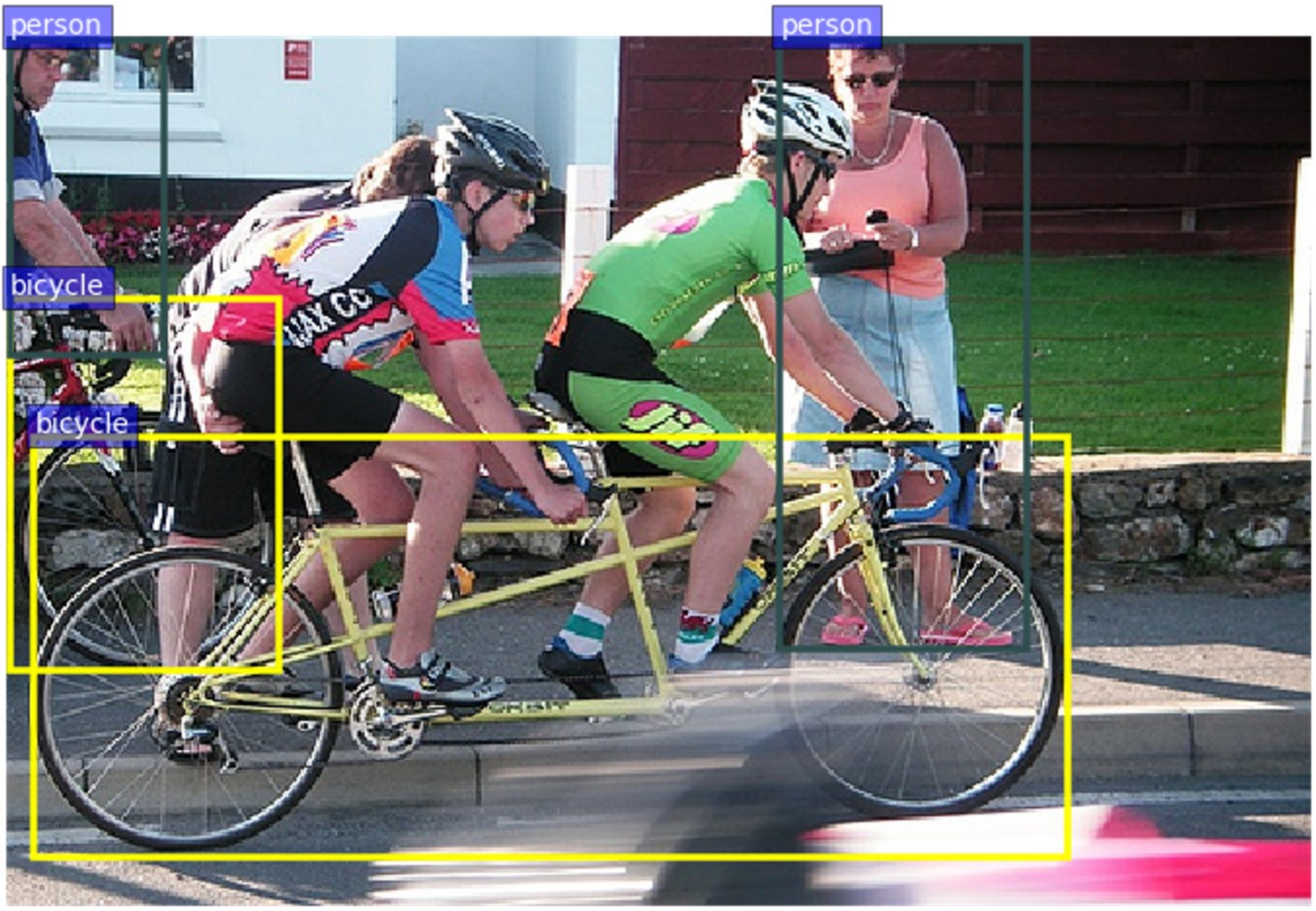}\hfill
	\includegraphics[width=0.235\textwidth]{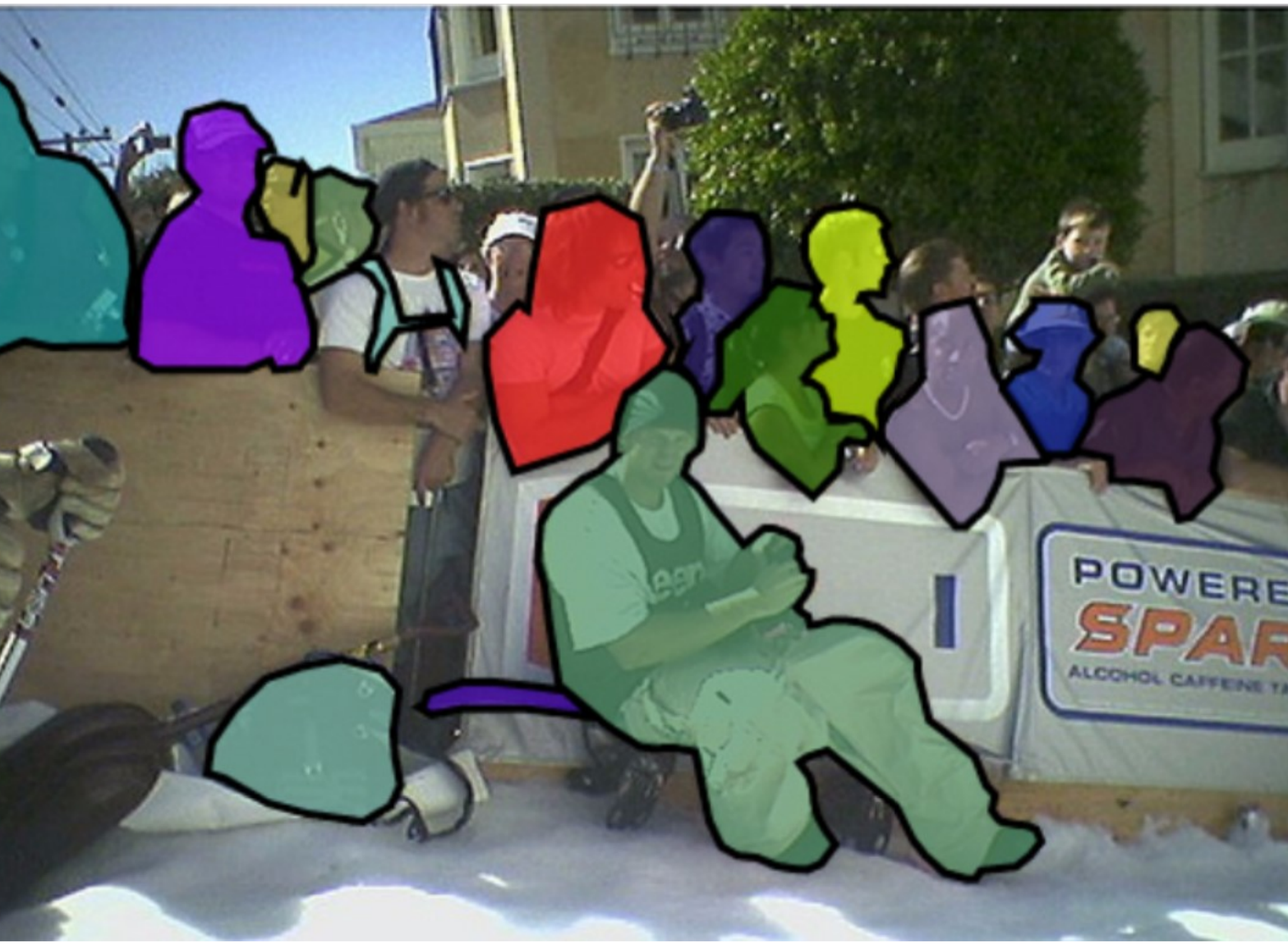}\hfill
	\includegraphics[width=0.235\textwidth]{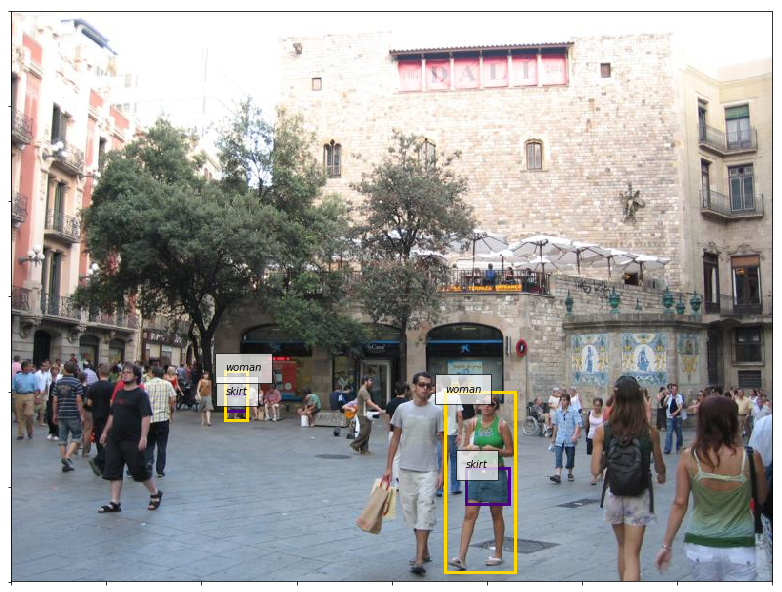}\hfill
	\includegraphics[width=0.23\textwidth]{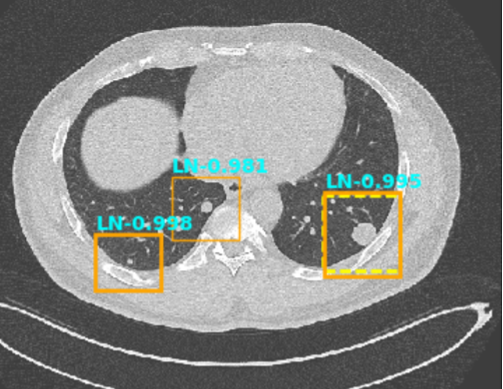}\hfill
	\caption{Because of the inherent difficulty of instance labeling, the ground truth of object detection datasets can be incomplete. Example images and their ground truth labels shown for (left to right) PASCAL VOC~\cite{Everingham2010} (missing people, bottles), MS COCO~\cite{Lin2014} (missing people), Visual Genome~\cite{Krishna2017} (missing people, tree, clothing, etc.), and DeepLesion~\cite{Yan2017} (ground truth is the dotted line; two boxes on the left indicate two unlabeled nodules).}
	\label{fig:missing_ex}
\end{figure}

The performance of supervised deep learning models is often highly dependent on the quality of the labels they are trained on \cite{Zhang2017,Veit2017,Jiang2018}.
Recent work by \cite{Toneva2019} has implied the existence of ``support vectors'' in datasets: hard to classify examples that have an especially significant influence on a classifier's decision boundary.
As such, ensuring that these difficult examples have the correct label would appear to be important to the final classifier.
However, collecting complete labels for object detection \cite{Girshick2014,Girshick2015,Liu2015,Ren2015,Redmon2016,Dai2016,Redmon2017} can be challenging, much more so than for classification tasks.
Unlike the latter, with a single label per image, the number of objects in an image is often variable, and objects can come in a large variety of shapes, sizes, poses, and settings, even within the same class.
Furthermore, object detection scenes are often crowded, resulting in object instances that may be occluded.
Given the requirement for tight bounding boxes and the sheer number of instances to label, constituting annotations can be very time-consuming.
For example, just labeling instances, without localization, required $\sim$30K worker hours for the 328K images of MS COCO \cite{Lin2014}, and the airport checkpoint X-ray dataset used in \cite{Liang2018a,Sigman2020}, which required assembling bags, scanning, and hand labeling, took over 250 person hours for 4000 scans over the span of several months.
For medical datasets \cite{Moreira2012,Yan2017,Lee2017,Wang2017}, this becomes even more problematic, as highly trained (and expensive) radiologist experts or invasive biopsies are needed to determine ground truth.

As a result of its time-consuming nature, dataset annotations are often crowd-sourced when possible, either with specialized domain experts or Amazon Mechanical Turk.
In order to ensure consistency, dataset designers establish labeling guidelines or have multiple workers label the same image \cite{Everingham2010,Lin2014}.
Regardless, tough judgment calls, inter- and even intra-worker variability, and human error can still result in overall inconsistency in labeling, or missing instances entirely.
This becomes especially exacerbated when establishing a larger dataset, like OpenImages \cite{Kuznetsova2018}, which while extremely large, is incompletely labeled.

Despite this, object detection algorithms often use the standard cross-entropy loss for object classification.
Implicit to this loss function is the assumption that any region without a bounding box does not contain an object; in other words, classification is posed as a positive-negative (PN) learning problem.
While reasonable for a completely labeled dataset, despite best efforts, this is often not the case in practice due to the previously outlined difficulties of instance annotations.
As shown in Figure \ref{fig:missing_ex} for a wide array of common datasets, the lack of instance labels does not always mean the absence of true objects.

While the result of this characterization constitutes a noisy label setting, it is not noisy in the same respect as is commonly considered for classification problems \cite{Zhang2017,Veit2017,Jiang2018}.
The presence of a positive label in object detection datasets are generally correct with high probability; it is the \textit{lack} of a label that should not be interpreted with confidence as a negative (or background) region.
Thus, given these characteristics common to object detection data, we propose recasting object detection as a positive-unlabeled (PU) learning problem \cite{Denis1998,DeComite1999,Letouzey2000,Elkan2008,Kiryo2017}.
With such a perspective, existing labels still implies a positive sample, but the lack of one no longer enforces that the region must be negative.
This can mitigate the confusing learning signal that often occurs when training on object detection datasets.

In this work, we explore how the characteristics of object detection annotation lend themselves to a PU learning problem and demonstrate the efficacy of adapting detection model training objectives accordingly.
We perform a series of experiments to demonstrate the effectiveness of the PU objective on two popular, well-labeled object detection datasets (PASCAL VOC \cite{Everingham2010} and MS COCO \cite{Lin2014}) across a range of label missingness, as well as two datasets with real incomplete labels (Visual Genome \cite{Krishna2017}, DeepLesion \cite{Yan2017}). 

\section{Methods}
\begin{figure*}[t]
	\centering
	\subfigure[Positive-Negative, with full labels]{\centering\includegraphics[scale=0.6,width=.25\textwidth]{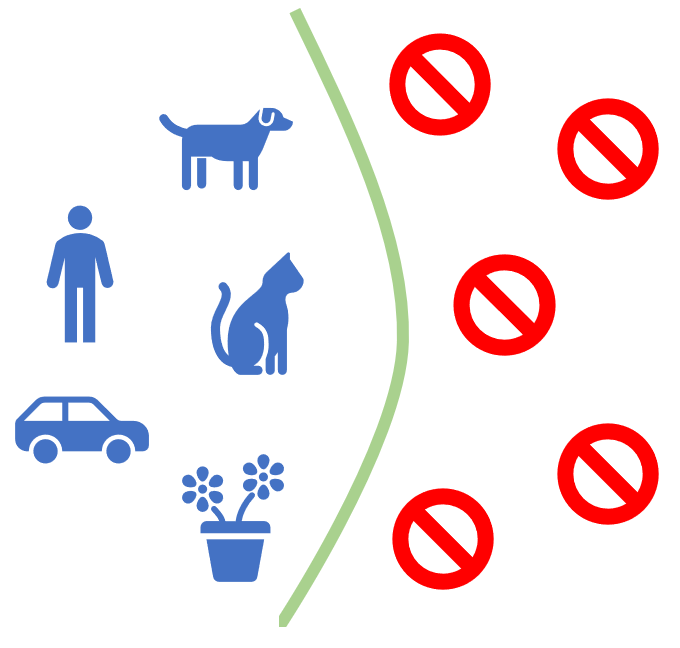}}\qquad
	\subfigure[Positive-Negative, with missing labels]{\centering\includegraphics[scale=0.6,width=.25\textwidth]{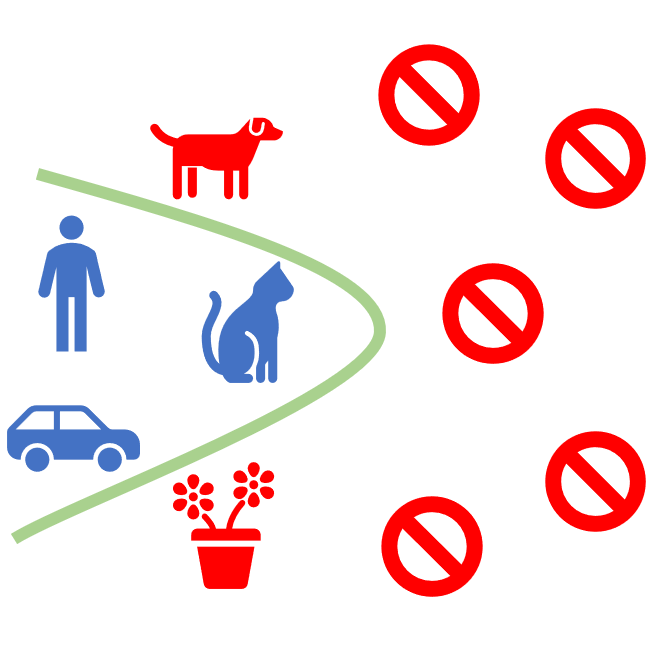}}\qquad
	\subfigure[Positive-Unlabeled]{\centering\includegraphics[scale=0.6,width=.25\textwidth]{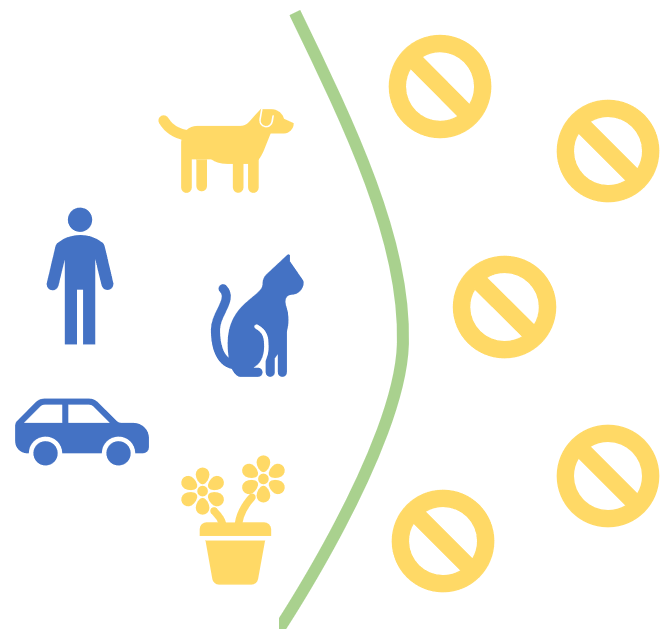}}
	\caption{A classifier (green) learns to separate proposals by ``objectness''. Models trained with a standard cross-entropy loss implicitly assume positive-negative (PN) learning: regions with bounding boxes are considered positive (blue), and any other proposed boxes are treated as negative (red). This is reasonable when labels are complete (a), but in reality, object detection datasets are inherently missing annotations; this forces the classifier to exclude unlabeled objects from the positive class (b). We propose a positive-unlabeled (PU) approach (c), which considers non-labeled regions as unlabeled (yellow) rather than negative, allowing non-positive regions to be classified as positive. Best viewed in color.}
	\label{fig:PN_v_PU_learning}
\end{figure*}

\subsection{Faster R-CNN}\label{sec:FRCNN}
In principle, the observed problem is characteristic of the data and is thus general to any object detection framework.
However, in this work, we primarily focus on Faster R-CNN~\cite{Ren2015}, a popular 2-stage method for which we provide a quick overview here.

As with other object detection models, given an input image $X$, the desired output of Faster R-CNN is a bounding box $B^{(i)} \in \mathbb R^4$ and class probabilities $c^{(i)} \in \mathbb R^k$ for each object (indexed by $i$) present, where $k$ is the number of classes and the final classification decision is commonly $\argmax c^{(i)}$.
Faster R-CNN does this in a 2-stage process.
First, a convolutional neural network (CNN)~\cite{Lecun1989} is used to produce image features $h$.
A Region Proposal Network (RPN) then generates bounding box proposals $\hat{B}^{(i)}$ relative to a set of reference boxes spatially tiled over $h$.
At the same time, the RPN predicts an ``objectness'' probability $\hat{c}^{(i)}$ for each proposal, learned as an object-or-not binary classifier.
The second stage then takes the proposals with the highest scores, and predicts bounding box refinements to produce $B^{(i)}$ and the final classification probabilities $c^{(i)}$.

Of particular interest is how the classifier producing $\hat{c}^{(i)}$ is trained.
Specifically, the cross-entropy loss $H(t,y)$ is employed, where $H(t,y)$ signifies the loss incurred when the model outputs $t$ when the ground truth is $y$.
In the RPN, this results in the following classification risk minimization:
\begin{equation} \label{eq:RPN_risk}
R^{RPN}_{pn} = \pi_p \mathbb{E}[H(\hat{c}_p,+1)] +  \pi_n \mathbb{E}[H(\hat{c}_n,-1)]
\end{equation}
where $\pi_p$ and $\pi_n$ are the class probability priors for the positive and negative classes, respectively, and $\hat{c}^{(i)}_p$ and $\hat{c}^{(i)}_n$ are the predicted ``objectness'' probabilities for ground truth positive and negative regions.
This risk is estimated with samples as:
\begin{equation} \label{eq:RPN_loss}
\mathcal{L}^{RPN}_{pn} = \frac{\hat{\pi}_p}{N_p} \sum_{i=1}^{N_p} H(\hat{c}^{(i)}_p,+1) + \frac{\hat{\pi}_n}{N_n} \sum_{i=1}^{N_n} H(\hat{c}^{(i)}_n,-1) 
\end{equation}
where $N_p$ and $N_n$ are the number of ground truth positive and negative regions being considered, respectively, and the class priors are typically estimated as $\hat{\pi}_p=\frac{N_p}{N_p + N_n}$ and $\hat{\pi}_n=\frac{N_n}{N_p + N_n}$.
Notably, this training loss treats all non-positive regions in an image as negative.

\subsection{PU learning}
In a typical binary classification problem, input data $X\in\mathbb{R}^d$ are labeled as $Y\in\{\pm1\}$, resulting in what is commonly termed a positive-negative (PN) problem. 
This implicitly assumes having samples from both the positive (P) and negative (N) distributions, and that these samples are labeled correctly (Figure \ref{fig:PN_v_PU_learning}a).
However, in some scenarios, we only have labels for some positive samples.
The remainder of our data are \textit{unlabeled} (U): samples that could be positive or negative.
Such a situation is called a positive-unlabeled (PU) setting, where the N distribution is replaced by an unlabeled (U) distribution (Figure \ref{fig:PN_v_PU_learning}c).
Such a representation admits a classifier that can appropriately include unlabeled positive regions on the correct side of the decision boundary.
We briefly review PN and PU risk estimation here.

Let $p(x,y)$ be the underlying joint distribution of $(X,Y)$, $p_p(x)=p(x|Y=+1)$ and $p_n(x)=p(x|Y=-1)$ be the distributions of P and N data, $p(x)$ be the distribution of U data, $\pi_p=p(Y=+1)$ be the positive class-prior probability, and $\pi_n=p(Y=-1)=1-\pi_p$ be the negative class-prior probability. 
In a PN setting, data are sampled from $p_p(x)$ and $p_n(x)$ such that $\mathcal X_p=\{x^p_i\}^{N_p}_{i=1}\sim{p_p(x)}$ and $\mathcal X_n=\{x^n_i\}^{N_n}_{i=1}\sim{p_n(x)}$.
Let $g$ be an arbitrary decision function that represents a model.
The risk of $g$ can be estimated from $\mathcal X_p$ and $\mathcal X_n$ as:

\begin{equation} \label{eq:PN_loss}
\hat{R}_{pn}(g)=\pi_p\hat{R}^{+}_p(g)+\pi_n\hat{R}^{-}_n(g)
\end{equation}
$\hat{R}^{+}_p(g)=1/N_p\sum^{N_p}_{i=1}\ell(g(x^p_i),+1)$ and $\hat{R}^{-}_n(g)=1/N_n\sum^{N_n}_{i=1}\ell(g(x^n_i),-1)$, where $\ell$ is the loss function. 
In classification, $\ell$ is commonly the cross-entropy loss $H(t,y)$.

In PU learning, $\mathcal X_n$ is unavailable; instead we have unlabeled data $\mathcal X_u=\{x^u_i\}^{N_u}_{i=1}\sim{p(x)}$, where $N_u$ is the number of unlabeled samples.
However, the negative class empirical risk $\hat{R}^{-}_n(g)$ in Equation \ref{eq:PN_loss} can be approximated indirectly~\cite{du2015convex, du2014analysis}. 
Denoting $R^{-}_p(g)=\mathbb{E}_p[\ell(g(X),-1)]$ and $R^{-}_u(g)=\mathbb{E}_{X\sim{p(x)}}[\ell(g(X),-1)]$, and observing $\pi_np_n(x)=p(x)-\pi_pp_p(x)$, we can replace the missing term $\pi_n R_n^-(g)=R^-_u(g)-\pi_pR^-_p(g)$. 
Hence, we express the overall risk without explicit negative data as
\begin{equation} \label{eq:PU_loss}
\hat{R}_{pu}(g)=\pi_p\hat{R}^{+}_p(g)+\hat{R}^-_u(g)-\pi_p\hat{R}^{-}_p(g)
\end{equation}
where $\hat{R}^{-}_p(g)=1/N_p\sum^{N_p}_{i=1}\ell(g(x^p_i),-1)$ and $\hat{R}^{-}_u(g)=1/N_u\sum^{N_u}_{i=1}\ell(g(x^u_i),-1)$. 

However, a flexible enough model can overfit the data, leading to the empirical risk in Equation \ref{eq:PU_loss} becoming negative.
Given that most modern object detectors utilize neural networks, this type of overfitting can pose a significant problem.
In \cite{Kiryo2017}, the authors propose a non-negative PU risk estimator to combat this:
\begin{equation} \label{eq:non-negative_PU_loss}
\hat{R}_{pu}(g)=\pi_p\hat{R}^{+}_p(g)+\max\{0,\hat{R}^-_u(g)-\pi_p\hat{R}^{-}_p(g)\}
\end{equation}
We choose to employ this non-negative PU risk estimator for the rest of this work.

\subsection{PU learning for object detection} \label{sec:PU-RPN}
\subsubsection{PU object proposals}
In object detection datasets, the ground truth labels represent positive samples.
Any regions that do not share sufficient overlap with a ground truth bounding box are typically considered as negative background, but the accuracy of this assumption depends on every object within a training image being labeled, which may not be the case (Figure~\ref{fig:missing_ex}). 
As shown in Figure~\ref{fig:PN_v_PU_learning}b, this results in the possibility of positive regions being proposed that are labeled negative during training, due to a missing ground truth label; effects of this phenomenon during training are investigated empirically in Supplementary Materials Section \ref{apx:forgetting}.
We posit that object detection more naturally resembles a PU learning problem than PN.

We recognize that two-stage detection naturally contains a binary classification problem in the first stage.
In Faster R-CNN specifically, the RPN assigns an ``objectness'' score, which is learned with a binary cross-entropy loss (Equation \ref{eq:RPN_loss}).
As previously noted, the PN nature of this loss can be problematic, so we propose replacing it with a PU formulation.
Combining Equations \ref{eq:RPN_loss} and \ref{eq:non-negative_PU_loss}, we produce the following loss function:
\begin{equation} \label{eq:PU-RPN}
\mathcal{L}^{RPN}_{pu} = \frac{\pi_p}{N_p} \sum_{i=1}^{N_p} H(\hat{c}^{(i)}_p,+1) +\max \bigg\{ 0, \frac{1}{N_u} \sum_{i=1}^{N_u} H(\hat{c}^{(i)}_u,-1) - \frac{\pi_p}{N_p} \sum_{i=1}^{N_p} H(\hat{c}^{(i)}_p,-1) \bigg\}
\end{equation}
Such a loss function relaxes the penalty of positive predictions for unlabeled objects.

\subsubsection{Estimating $\pi_p$} \label{sec:Est_pi_p}
\begin{wrapfigure}{r}{.4\linewidth}
    \vspace{-8mm}
	\centering
	\includegraphics[width=0.34\textwidth]{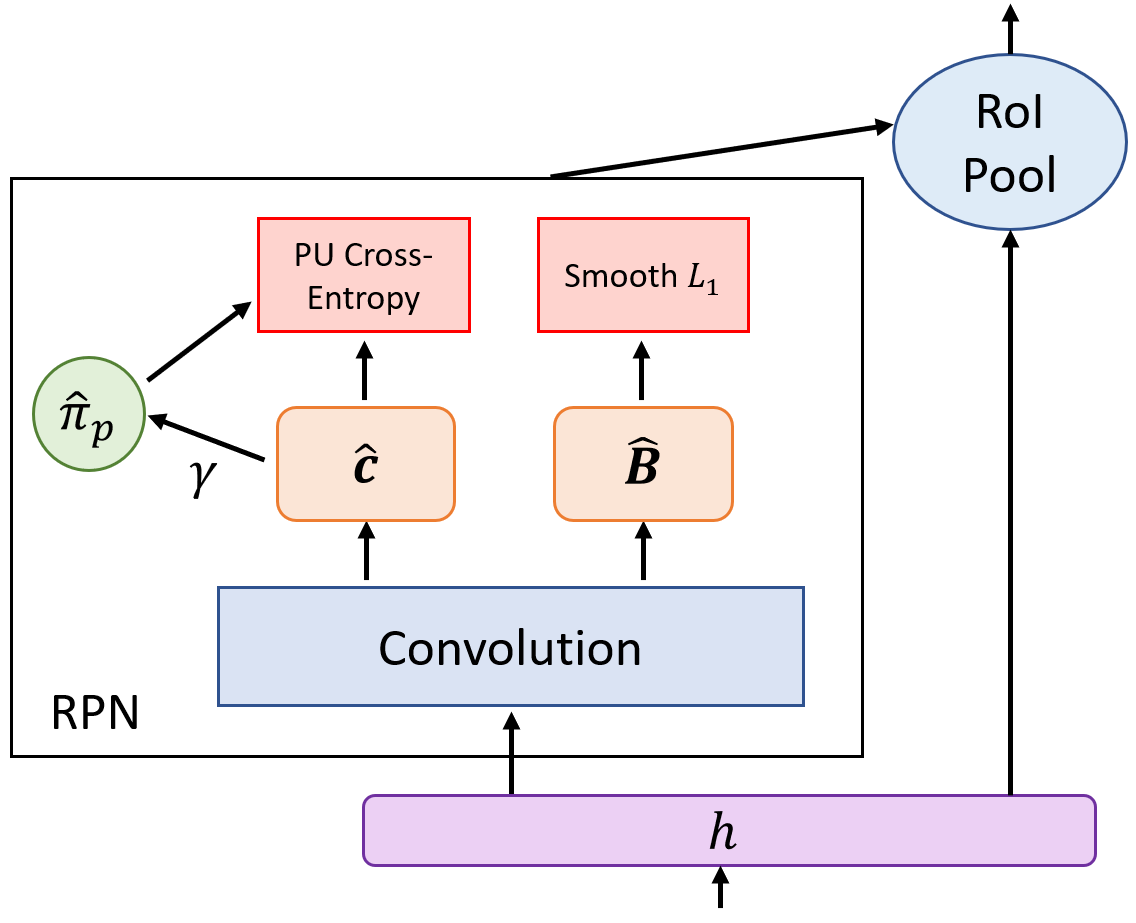}
	\caption{Faster R-CNN \cite{Ren2015} RPN with the proposed PU cross-entropy loss. Positive class prior estimate $\hat{\pi}_p$ is updated with objectness predictions $\hat{c}$, with momentum $\gamma$.}
	\label{fig:PU-RPN}
\end{wrapfigure}
The PU cross-entropy loss in Equation~\ref{eq:PU-RPN} assumes the class-prior probability of the positive class $\pi_p$ is known.
In practice, this is not usually the case, so $\pi_p$ must be estimated, denoted as $\hat{\pi}_p$.
For object detection, estimating $\pi_p$ is especially problematic because $\pi_p$ is not static: as the RPN is trained, an increasing proportion of region proposals will (hopefully) be positive.
While \cite{Kiryo2017} showed some robustness to $\pi_p$ misspecification, this was only on a fairly narrow range of $\pi_p \in [0.8 \pi_p, 1.2 \pi_p]$.
During object detection performance, $\pi_p$ starts from virtually $0$ and grows steadily as the RPN improves.
As such, any single estimate $\hat{\pi}_p$ poses the risk of being significantly off the mark during a large portion of training.

To address this, we recognize that the RPN of Faster R-CNN is already designed to infer the positive regions of an image, so we count the number of positive regions produced by the RPN and use it as an estimator for $\pi_p$:
\begin{equation} \label{eq:inf_pi_p}
\hat{\pi}_p = \frac{N_p^{RPN}}{N^{RPN}}
\end{equation}
where $N^{RPN}$ is the total number of RPN proposals that are sampled for training, and $N_p^{RPN}$ being those with classifier confidence of at least 0.5.
Note that this estimation of $\pi_p$ comes essentially for free.
Given that Faster R-CNN is trained one image at a time and the prevalence of objects varies between images, we maintain an exponential moving average with momentum $\gamma$ in order to stabilize $\hat{\pi}_p$ (see Figure \ref{fig:PU-RPN}).
This estimate $\hat{\pi}_p$ is then used in the calculation of the loss $\mathcal{L}^{RPN}_{pu}$ and its gradients.

\section{Related work}
Most object detection frameworks are designed fully supervised~\cite{Girshick2014,Girshick2015,Liu2015,Ren2015,Redmon2016,Dai2016,Redmon2017}: it is assumed that there exists a dataset of images where every object is labeled and such a dataset is available to train the model.
However, as discussed above, collecting such a dataset can be expensive.
Because of this, methods that can learn from partially labeled datasets have been a topic of interest for object detection.
What ``partially labeled'' constitutes can vary, and many types of label missingness have been considered.

Weakly supervised object detection models~\cite{Bilen2015,Oquab2015,Rochan2015} assume a dataset with image level labels, but without instance labels.
These models are somewhat surprisingly competent at identifying approximate locations of objects in an image without any object-specific cues, but have a harder time with precise localization.
This is especially the case when there are many of the same class of object in close proximity to each other, as individual activations can blur together, and the lack of bounding boxes makes it difficult to learn precise boundaries.
Other approaches consider settings where bounding boxes are available for some classes (e.g., PASCAL VOC's 20 classes) but not others (e.g., ImageNet~\cite{Deng2009} classes).
LSDA~\cite{Hoffman2014} does this by modifying the final layer of a CNN~\cite{Krizhevsky2012} to recognize classes from both categories, and \cite{Tang2016} improves upon LSDA by taking advantage of visual and semantic similarities between classes.
OMNIA~\cite{Rame2018} proposes a method merging datasets that are each fully annotated for their own set of classes, but not each other's.

There are also approaches that consider a single dataset, but the labels are undercomplete across all classes.
This setting most resembles what we consider in our paper.
In \cite{Dong2017}, only 3-4 annotated examples per class are assumed given to start; additional pseudo-labels are generated from the model on progressively more difficult examples as the model improves.
Soft-sampling has also been proposed to re-weight gradients of background regions that either have overlap with positive regions or produce high detection scores in a separately trained detector~\cite{Wu2018}; experiments were done on PASCAL VOC with a percentage of annotations discarded and on a subset of OpenImages~\cite{Kuznetsova2018}.

Positive-unlabeled learning has been proposed to improve statistical models with limited labeled data \cite{lee2003learning,Letouzey2000}. 
\cite{Kiryo2017,xu2017multi} analyzed the theoretical effectiveness of positive-unlabeled risk estimation on binary classification tasks, though with the proportion of positive class examples preset and known to the model. 
\cite{jain2016estimating,du2014analysis} infer the positive class prior with additional classifiers or matching algorithms. 
Our work extends the application of positive-unlabeled learning to alleviate the incomplete labels that can be found in most object detection datasets.

\section{Experiments}
For our experiments, we use the original Faster-RCNN \cite{Ren2015}, with a ResNet101~\cite{He2016} feature extractor (without feature pyramid network~\cite{lin2017feature}) pre-trained on ImageNet \cite{Deng2009}. 
To control the level of missingness of each dataset, each annotation from a object class provided by original dataset is randomly discarded with a probability $\rho$. 
This effectively makes the available annotations to be $1-\rho$ of total annotations on average.  

\subsection{Hand-tuning versus estimation of $\pi_p$}
\begin{figure}[t]
	\centering
	\subfigure[$\rho = 0.4$]{\centering\includegraphics[width=.33\textwidth]{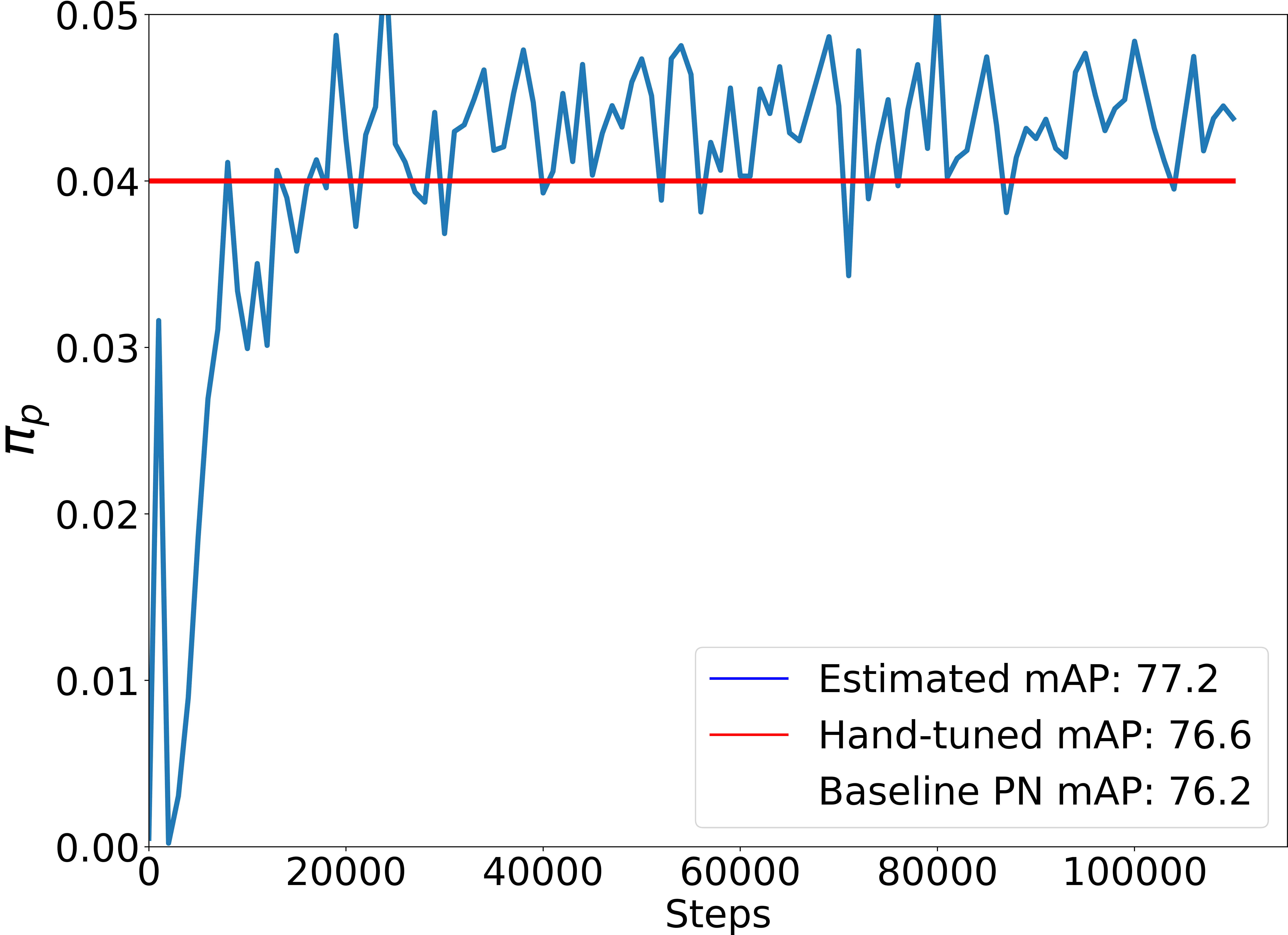}}\hfill
	\subfigure[$\rho = 0.5$]{\centering\includegraphics[width=.33\textwidth]{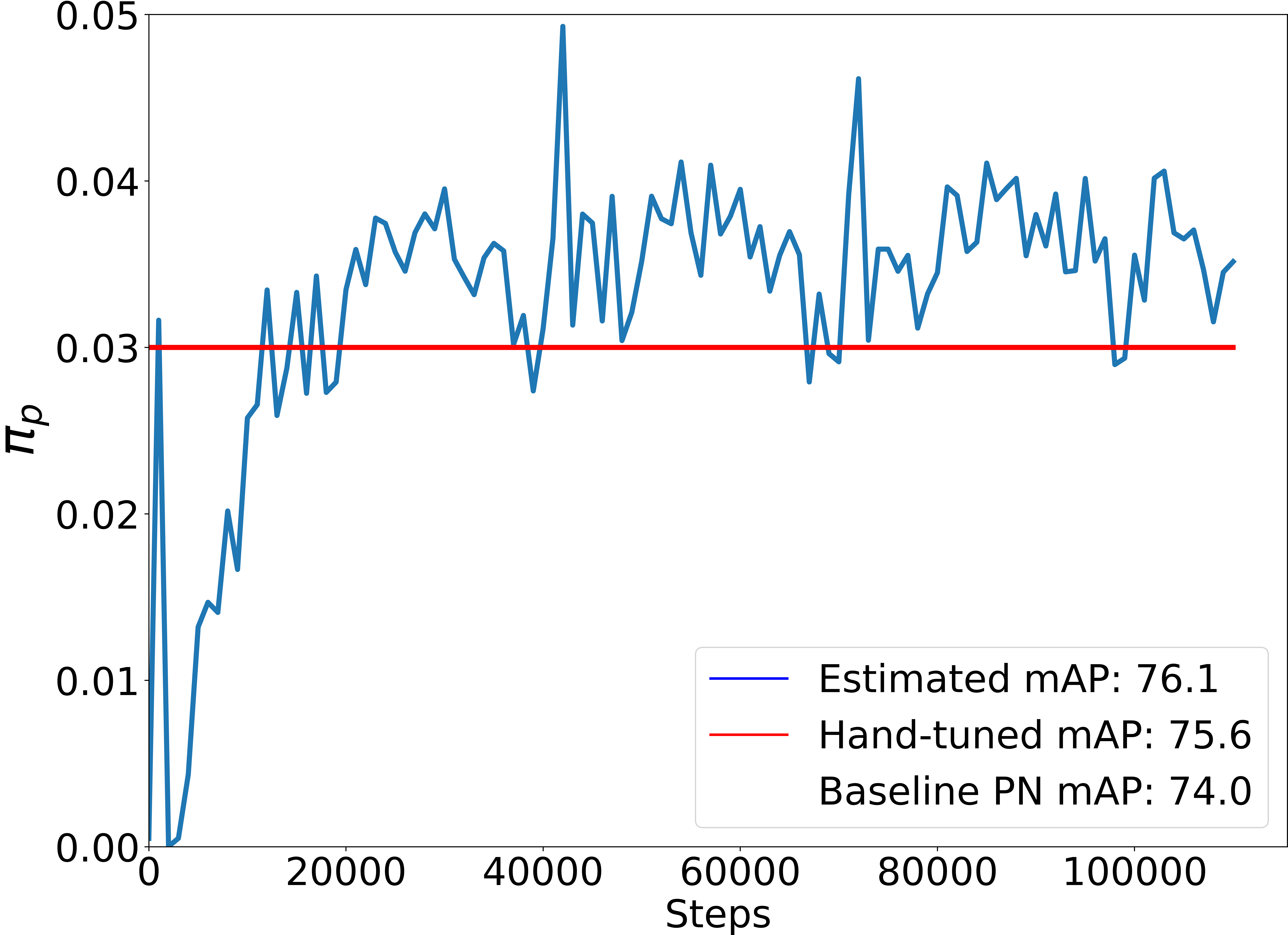}}\hfill
	\subfigure[$\rho = 0.6$]{\centering\includegraphics[width=.33\textwidth]{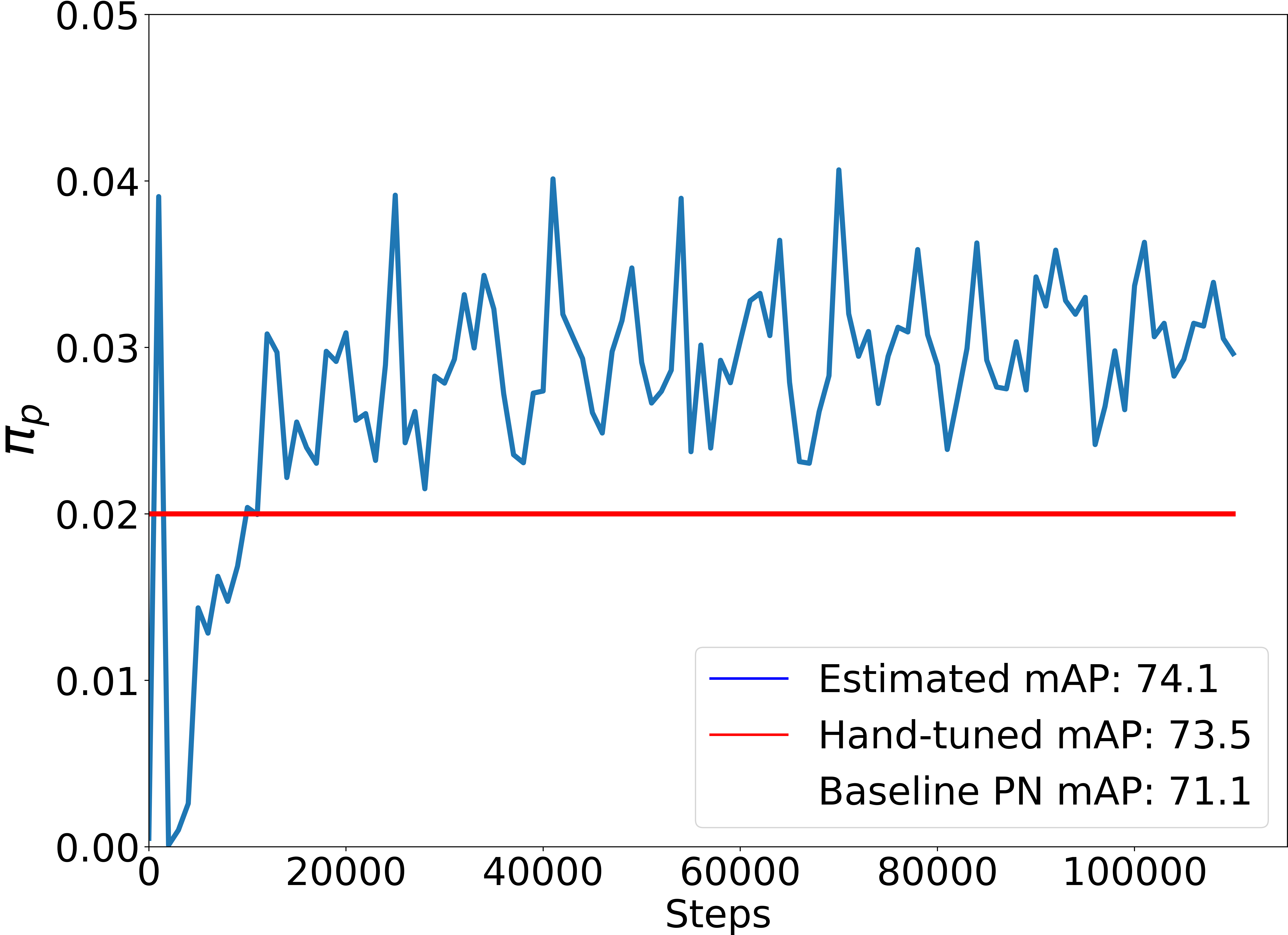}}\hfill
	\caption{Positive class prior $\hat{\pi}_p$ estimated during training of Faster R-CNN on PASCAL VOC versus from hand-tuning $\pi_p$ as a hyperparameter, for instance label missingness proportion $\rho = \{0.4, 0.5, 0.6\}$.}
	\label{fig:infer_pi}
\end{figure}

As discussed in Section \ref{sec:Est_pi_p}, PU risk estimation requires the prior $\pi_p$.
We experiment with two ways of determining $\pi_p$.
In the first method (\textit{Hand-Tuned}), we treat $\pi_p$ as a constant hyperparameter and tune it by hand.
In the second (\textit{Estimated}), we infer $\pi_p$ as described in Equation \ref{eq:inf_pi_p}, setting momentum $\gamma$ to 0.9.
We compare the estimate $\hat{\pi}_p$ inferred automatically with the hand-tuned $\pi_p$ that yielded the highest mAP on PASCAL VOC.
To see how our estimate changes in response to label missingness, when assembling our training set, we remove each annotation from an image with probability $\rho$, giving us a dataset with $1-\rho$ proportion of the total labels, and then do our comparison for $\rho = \{0.4, 0.5, 0.6\}$ in Figure \ref{fig:infer_pi}.

In all tested settings of $\rho$, the estimation $\hat{\pi}_p$ increases over time before stabilizing.
Such a result matches expectations, as when an object detection model is first initialized, its parameters have yet to learn good values, and thus the true proportion of positive regions $\pi_p$ is likely to be quite low.
As the model trains, its ability to generate accurate regions improves, resulting in a higher proportion of regions being positive. 
This in turn results in a higher true value of $\pi_p$, which our estimate $\hat{\pi}_p$ follows.
As the model converges, $\pi_p$ (and $\hat{\pi}_p$) stabilizes towards the true prevalence of objects in the dataset relative to background regions.
Interestingly, the final value of $\hat{\pi}_p$ settles close to the value of $\pi_p$ found by treating the positive class prior as a static hyperparameter, but consistently above it.
We hypothesize that this is due to a single static value having to hedge against the early stages of training, when $\pi_p$ is lower.
We use our proposed method of auto-inferring $\pi_p$ for the rest of our experiments, with $\gamma=0.9$, rather than hand-tuning it as a hyperparameter.
\begin{figure}[t]
	\centering
	\subfigure[PASCAL VOC]{\label{fig:exa}\includegraphics[width=.45\textwidth]{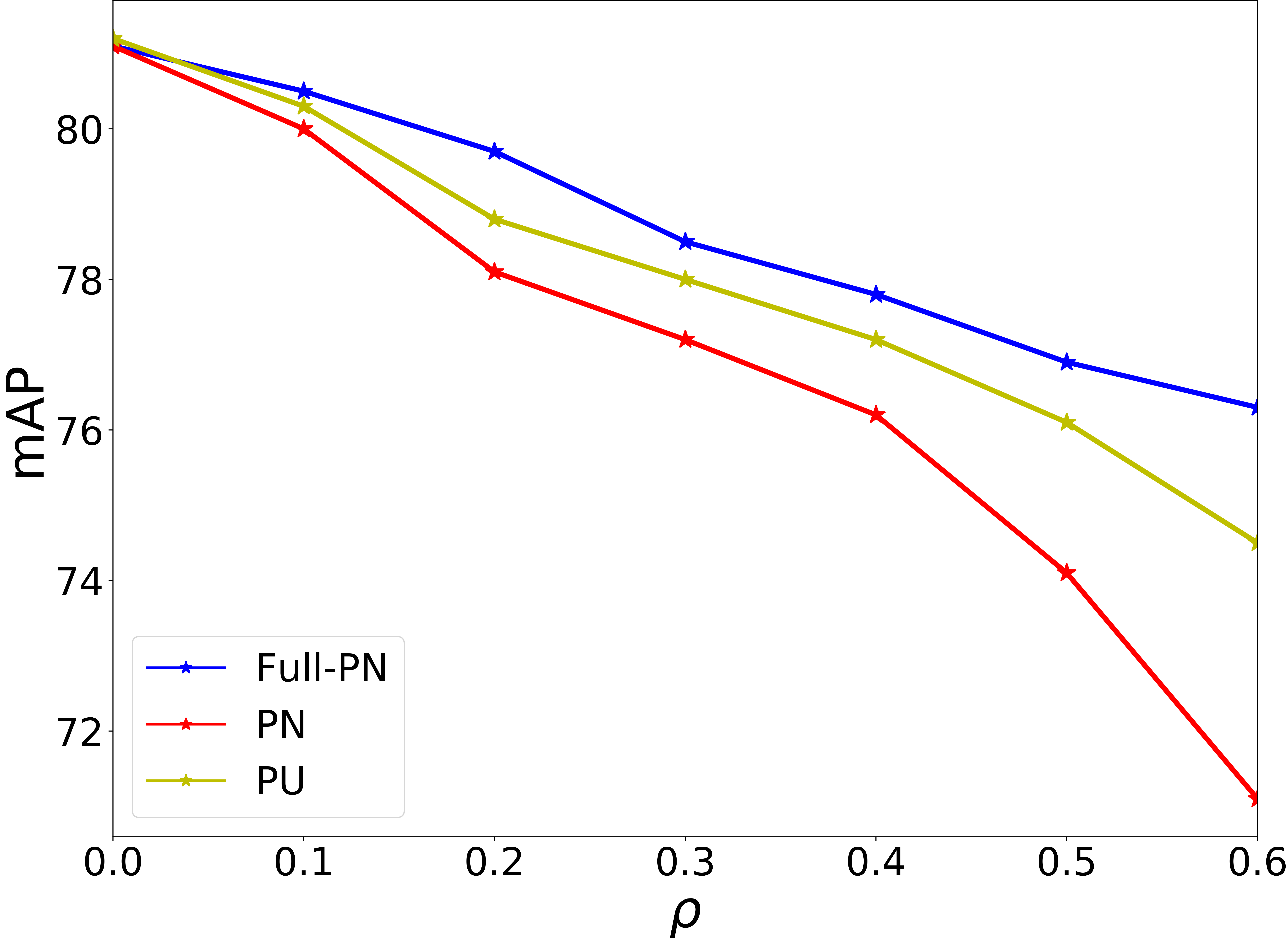}}
	\subfigure[MS COCO]{\label{fig:exb}\includegraphics[width=.45\textwidth]{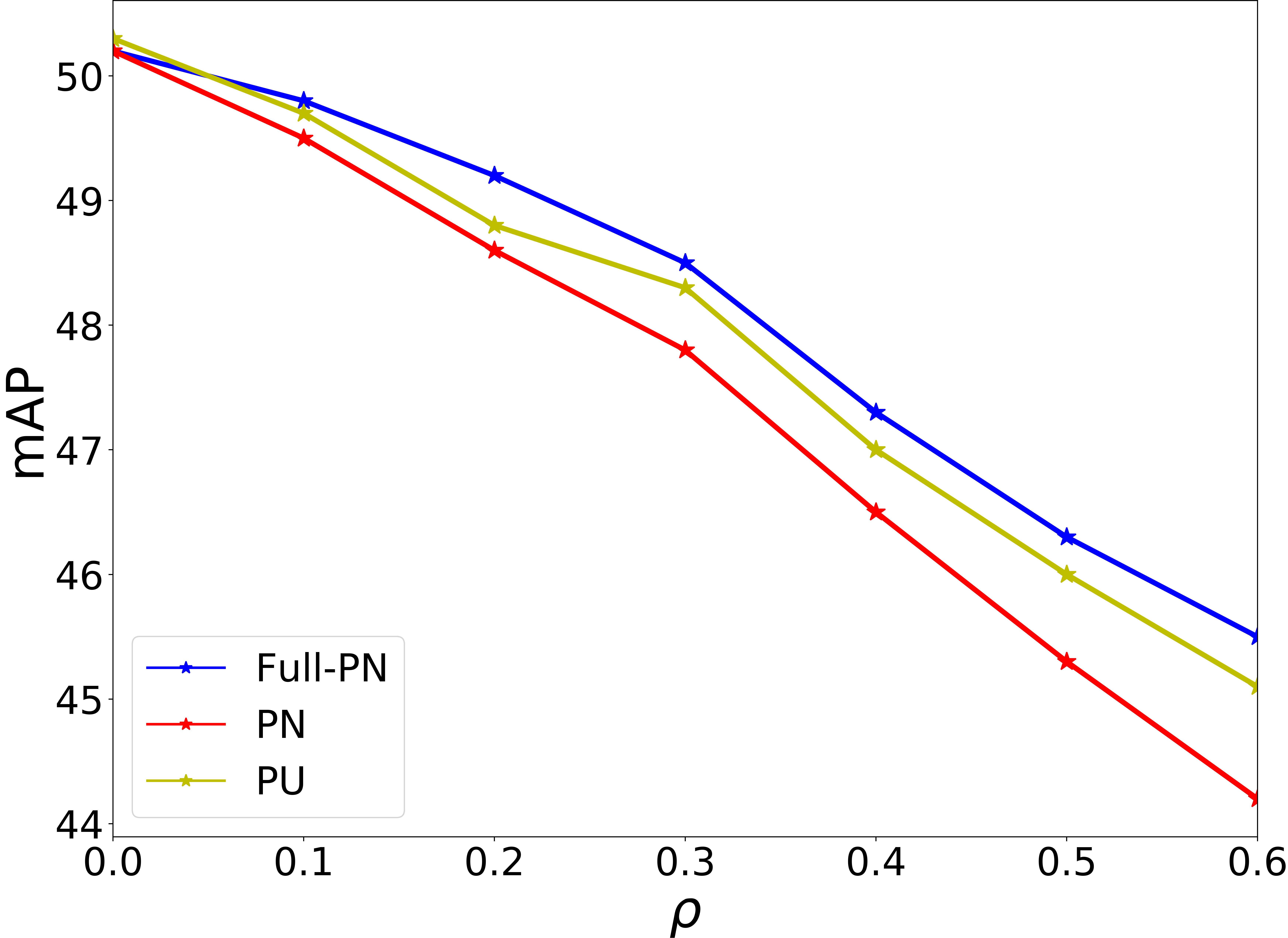}}
	\caption{mAP at IoU 0.5 ($\text{AP}_{50}$) on (a) PASCAL VOC and (b) MS COCO, for a range of label missingness $\rho$.}
	\label{PU_v_PN}	
\end{figure}

\subsection{PU versus PN on PASCAL VOC and MS COCO} \label{sec:PUvPN_VOC_COCO}
We investigate the effect that incomplete labels have on object detection training for the popular datasets PASCAL VOC \cite{Everingham2010} and MS COCO \cite{Lin2014}, using Faster R-CNN \cite{Ren2015} with a ResNet101 \cite{He2016} convolutional feature extractor.
In order to quantify the effect of missing labels, we artificially discard a proportion $\rho$ of the annotations.
We compare three settings, each for a range of values of $\rho$.
Given that the annotations are the source of the learning signal, we keep the number of total instances constant between settings for each $\rho$ as follows:

\begin{itemize}
	\item \textit{PN}: We remove a proportion of labels from every image in the dataset, such that the total proportion of removed labels is equal to $\rho$, and all images are included in the training set.
	We then train the detection model with a PN objective, as is normal.
	\vspace{-2mm}
	\item \textit{Full-PN}: We discard a proportion $\rho$ of entire images and their labels, resulting in a dataset of fewer images, but each of which retains its complete annotations.
	\vspace{-2mm}
	\item \textit{PU}: We use the same images and labels as in \textit{PN}, but instead train with our proposed PU objective.
\end{itemize}

A comparison of mean average precision (mAP) performance at IoU 0.5 for these 3 settings on PASCAL VOC and MS COCO is shown in Figure \ref{PU_v_PN}.
As expected, as $\rho$ is increased, the detector's performance degrades.
Focusing on the results for \textit{PN} and \textit{Full-PN}, it is clear that for an equal number of annotated objects, having fewer images that are more thoroughly annotated is preferable to a larger number of images with less thorough labels.
On the other hand, considering object detection as a PU (\textit{PU}) problem as we have proposed allows us to improve detector quality across a wide range of label missingness.
While having a more carefully annotated set (\textit{Full-PN}) is still superior, the PU objective helps close the gap.
Interestingly, there is a small gain (PASCAL VOC: +0.2, MS COCO: +0.3) in mAP at full labels ($\rho=0$), possibly due to better learning of objects missing labels in the full dataset.
Additional analysis on the improvements to RPN recall are in Supplemental Materials Section~\ref{apx:recall}.

\subsection{Visual Genome}
\begin{table}[b]
	\caption{Detector performance on Visual Genome, with full labels, at various IoU thresholds. 
	``Y'' indicates weighting by class frequency, while ``N'' denotes without weighting.}
	\begin{center}
		\resizebox{0.55\columnwidth}{!}{%
		\begin{tabular}{ l || c c | c c | c c }
			& \multicolumn{2}{c|}{ $\text{AP}_{25}$} & \multicolumn{2}{c|}{ $\text{AP}_{50}$} & \multicolumn{2}{c}{ $\text{AP}_{75}$} \\
			 & Y & N & Y & N & Y & N \\
			\hline
			\hline
			PN & 12.09 & 22.79 & 9.11 & 17.35 &	2.46 & 9.98 \\
			PU & \textbf{13.83} & \textbf{25.56} & \textbf{10.44} & \textbf{19.89} & \textbf{4.52} & \textbf{11.79} \\
		\end{tabular}
		}
	\end{center}
	\label{tab:VisualGenome}
\end{table}
Visual Genome~\cite{Krishna2017} is a scene understanding dataset of objects, attributes, and relationships.
While not as commonly used as an object detection benchmark as PASCAL VOC or MS COCO, Visual Genome is popular when relationships or attributes of objects are desired, as when Faster R-CNN is used as a pre-trained feature extractor for Visual Question Answering~\cite{Agrawal2015,Anderson2017}.
Given the large number of classes (33,877) and the focus on scene understanding during the annotation process, the label coverage of all object instances present in each image is correspondingly lower.
In order to achieve its scale, the labeling effort was crowd-sourced to a large number of human annotators.
As pointed out in \cite{Everingham2010}, even increasing from 10 classes of objects in PASCAL VOC2006 to the 20 in VOC2007 resulted in a substantially larger number of labeling errors, as it became more difficult for human annotators to remember all of the object classes.
This problem is worse by several orders of magnitude for Visual Genome.
While the dataset creators implemented certain measures to ensure quality, there still are many examples of missing labels.
In such a setting, the proposed PU risk estimation is especially appropriate, even with all included labels. 

We train ResNet101 Faster R-CNN using both PN and the proposed PU risk estimation on 1600 of the top object classes of Visual Genome, as in~\cite{Anderson2017}.
We evaluate performance on the classes present in the test set and report mAP at various IoU thresholds $\{0.25, 0.50, 0.75\}$ in Table~\ref{tab:VisualGenome}.
We also show mAP results when each class's average precision is weighted according to class frequency, as done in \cite{Anderson2017}.
The PASCAL VOC and MS COCO results in Figure~\ref{PU_v_PN} indicate that we might expect increasing benefit from utilizing a PU loss as missing labels become especially prevalent, and for Visual Genome, where this is indeed the case, we observe that PU risk estimation outperforms PN by a significant margin, across all settings.
Similar improvements are also observed on OpenImages \cite{Krasin2017,Kuznetsova2018}, a dataset with a similar degree of object missingness (Supplemental Materials Section~\ref{apx:openimages}).

\subsection{DeepLesion}
\begin{figure}[t]
	\centering
	\subfigure[]{\includegraphics[width=.45\textwidth]{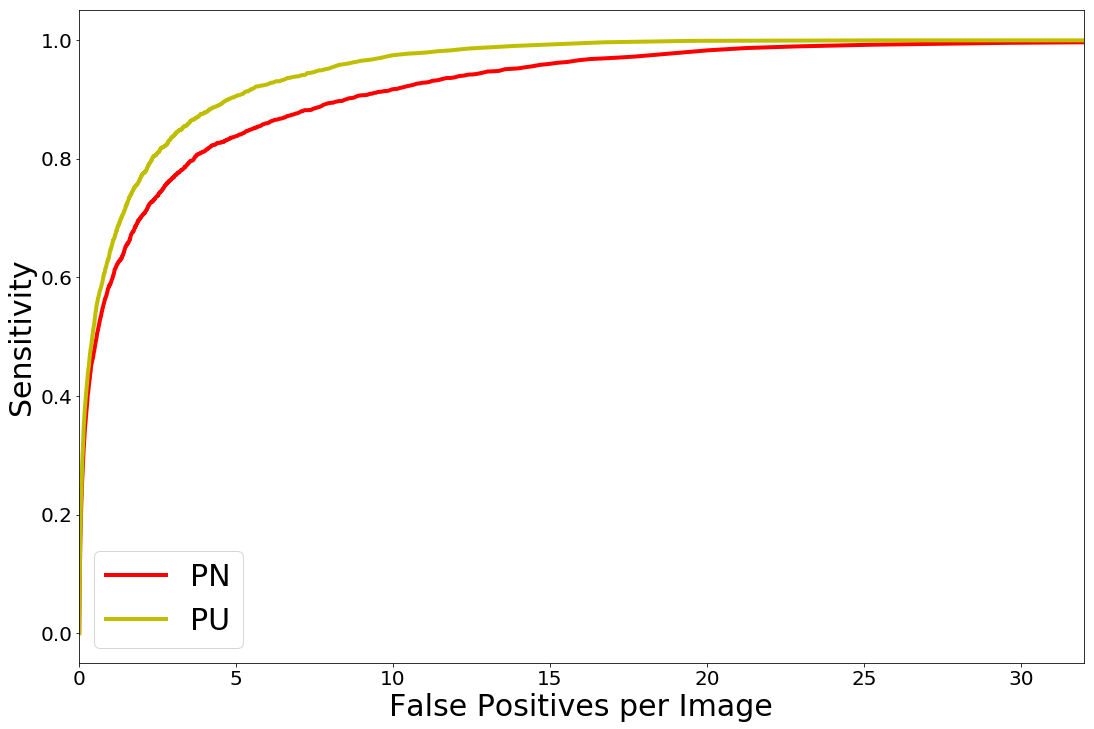}}
	\subfigure[]{\includegraphics[width=.45\textwidth]{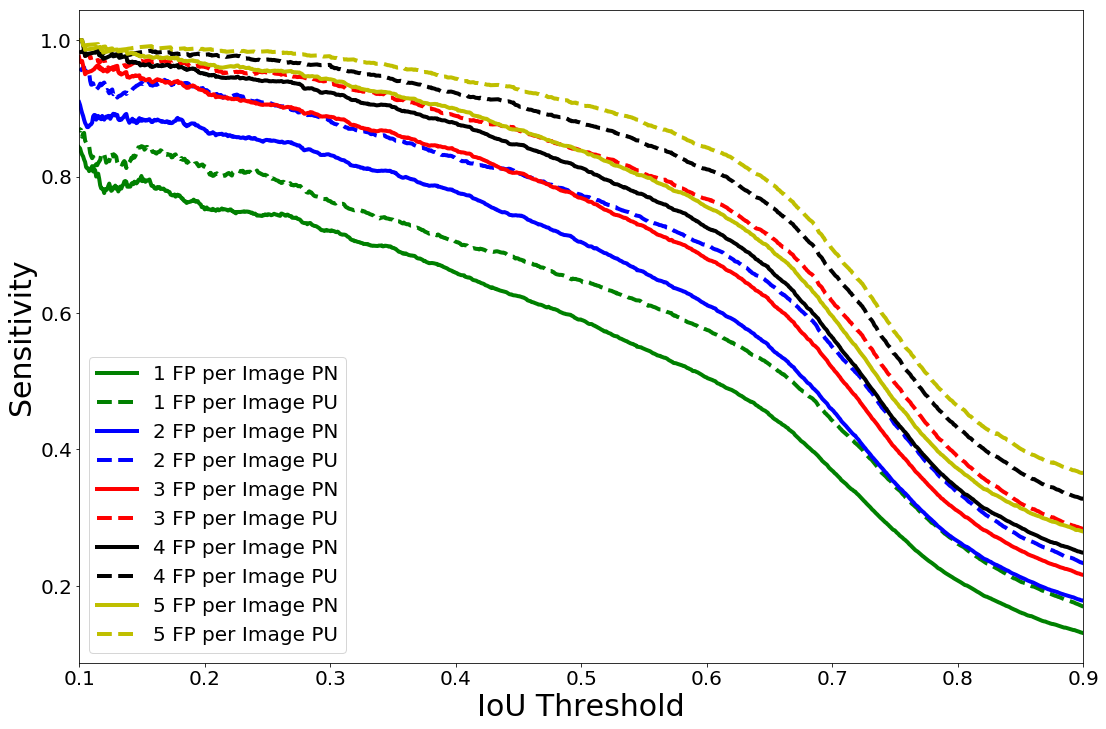}}
	\caption{Lesion sensitivity versus (a) false positive rate and (b) IoU threshold for different false positive (FP) allowances per image. We compare the baseline Faster R-CNN variant in \cite{Yan2017} trained with a PN objective versus the proposed PU objective.}
	\label{deep_lesion_sensitivity}	
\end{figure}
The recent progress in computer vision has attracted increasing attention towards potential health applications.
To encourage deep learning research in this direction, the National Institutes of Health (NIH) Clinical Center released DeepLesion \cite{Yan2017}, a dataset consisting of 32K CT scans with annotated lesions.
Unlike PASCAL VOC, MS COCO, or Visual Genome, labeling cannot be crowd-sourced for most medical datasets, as accurate labeling requires medical expertise.
Even with medical experts, labeling can be inconsistent; lesion detection is a challenging task, with biopsy often necessary to get an accurate result.
Like other datasets labeled by an ensemble of annotators, the ground truth of medical datasets may contain inconsistencies, with some doctors being more conservative or aggressive in their diagnoses.
Due to these considerations, a PU approach more accurately characterizes the nature of the data.

We re-implemented the modified version of Faster R-CNN described in \cite{Yan2017} as the baseline model and compare against our proposed model using the PU objective, making no other changes.
We split the dataset into 70\%-15\%-15\% parts for training, validation, and test.
Following \cite{Yan2017}, we report results in terms of free receiver operating characteristic (FROC) and sensitivity of lesion detection versus intersection-over-union (IoU) threshold for a range of allowed false positives (FP) per image (Figure \ref{deep_lesion_sensitivity}).
In both cases, we show that switching from a PN objective to a PU one results in gains in performance.

\section{Conclusion and future work}
Having observed that object detection data more closely resembles a positive-unlabeled (PU) problem, we propose training object detection models with a PU objective.
Such an objective requires estimation of the class probability of the positive class, but we demonstrate how this can be estimated dynamically with little modification to the existing architecture.
Making these changes allows us to achieve improved detection performance across a diverse set of datasets, some of which are real datasets with significant labeling difficulties.
While we primarily focused our attention on object detection, a number of other popular tasks share similar characteristics and could also benefit from being recast as PU learning problems (e.g., segmentation~\cite{Ronneberger2015,Long2015,He2017}, action detection~\cite{Soomro2012,Idree2017,Gu2018}).

In our current implementation, we primarily focus on applying the PU objective to the binary object-or-not classifier in Faster R-CNN's Region Proposal Network.
A natural extension of this work would be to apply the same objective to the second stage classifier, which must also separate objects from background.
However, as the second stage classifier outputs one of several classes (or background), the classification is no longer binary, and requires estimating multiple class priors $\{\pi_c\}_{c=1}^k$ \cite{xu2017multi}, which we leave to future work.
Such a multi-class PU loss would also allow extension to single-stage detectors like SSD~\cite{Liu2015} and YOLO~\cite{Redmon2016,Redmon2017}.
Given the performance gains already observed, we believe this to be an effective and natural improvement to the object detection classification loss.

\section*{Acknowledgments}
The work was funded in part by DARPA, DOE, NIH, NSF, ONR, and TSA.
The authors thank John Sigman and Gregory Spell for feedback on an early draft.

\bibliography{bib}

\newpage
\appendix
\section{Example forgetting in object detection}\label{apx:forgetting}
\vspace{-1mm}
\begin{figure}[h]
    \centering
	\includegraphics[width=0.9\textwidth]{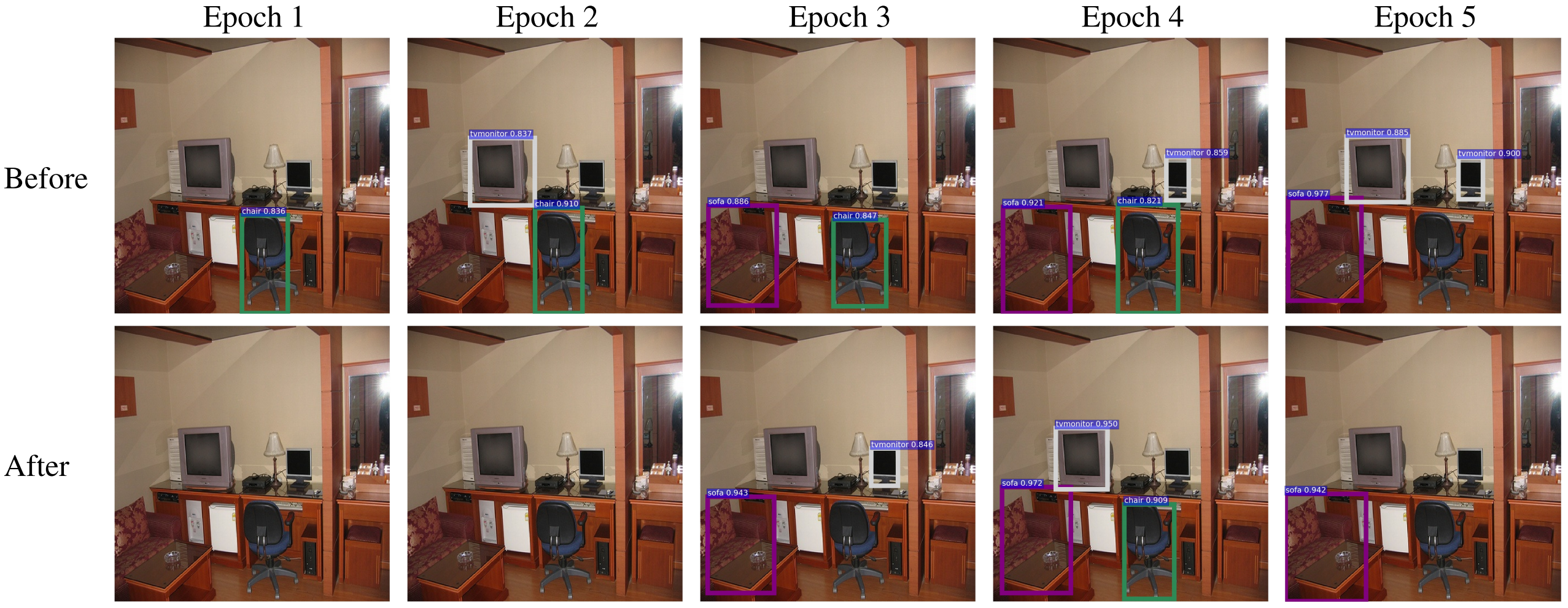}
	\caption{Detections on a PASCAL VOC train set image missing annotations throughout training: only the sofa in the lower left has a label. Each column shows detections directly before (top) and after (bottom) the model is trained on the image shown, for each epoch. While the sofa is consistently detected (purple box) after being learned, the unlabeled objects (2 monitors, a chair) are repeatedly found and then suppressed after being trained upon.} 
	\label{fig:forgetting_ex}
\end{figure}

In a recent study of training dynamics of neural network classifiers, \citet{Toneva2019} defined a ``forgetting event'' as a training example switching from being classified correctly by the model to being classified incorrectly during training. 
It was found that certain examples were forgotten more frequently than others while others were never forgotten (termed ``unforgettable''), with the degree of forgetting for individual examples being consistent across neural network architectures and random seeds.
When visualized, the forgotten examples tend to have atypical or uncommon characteristics (e.g., pose, lighting, angle), relative to ``unforgettable'' examples.
Interestingly, a significant number of ``unforgettable'' examples could be removed from the training set with only a marginal reduction in test accuracy, if the ``hard'' examples were kept.
This implies that the ``hard'' examples play a role akin to support vectors in max-margin learning, while easier ``unforgettable'' examples have little effect on the final decision boundary.

Within the context of object detection datasets, we hypothesize that unlabeled object instances form a similar group of hard examples that are also learned and then forgotten throughout training.
Unlike the inter-batch catastrophic forgetting in \cite{Toneva2019}, however, where hard examples are learned while part of the current minibatch and then forgotten while learning other examples, unlabeled samples in object detection are learned from other examples and then \textit{suppressed} after incurring misclassification losses during training (see Figure \ref{fig:forgetting_ex}).

Unlabeled instances strongly resemble positive examples throughout the rest of the dataset, but their lack of labels mean that the typical PN classification objective incentivizes learning them as negatives.
Given that hard examples have a strong influence on classifier boundaries, having unlabeled examples trained as negatives may prove especially detrimental to training.

We perform a similar study as \cite{Toneva2019} and investigate forgetting events on PASCAL VOC~\cite{Everingham2010} by tracking detection rates of labeled and unlabeled instances in the training set throughout learning.
In particular, an object is considered detected if the detector produces a bounding box with intersection over union (IoU) of at least 0.5 and the classifier is at least 80\% confident in the correct class. 
We track whether or not an object was detected directly before the image it belongs to is trained upon, and then again after the gradients have been applied.
These indicator variables are then combined across objects for each epoch and reported as a percentage.
While PASCAL VOC does naturally have unlabeled instances, we do not have access to these without a re-labeling effort.
As such, we remove $10\%$ of object annotations randomly across all object classes during training, and use them to calculate detection rates for this experiment.

\begin{figure*}[t]
	\centering
	\subfigure[Labeled objects]{\includegraphics[width=.40\textwidth]{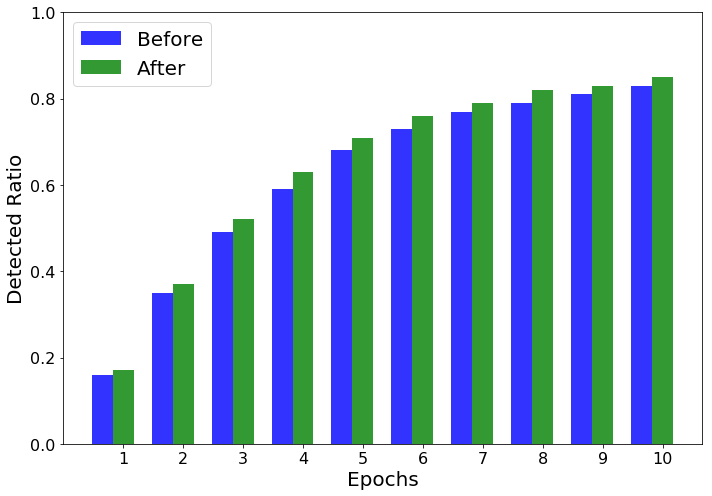}}\qquad
	\subfigure[Unlabeled objects]{\includegraphics[width=.40\textwidth]{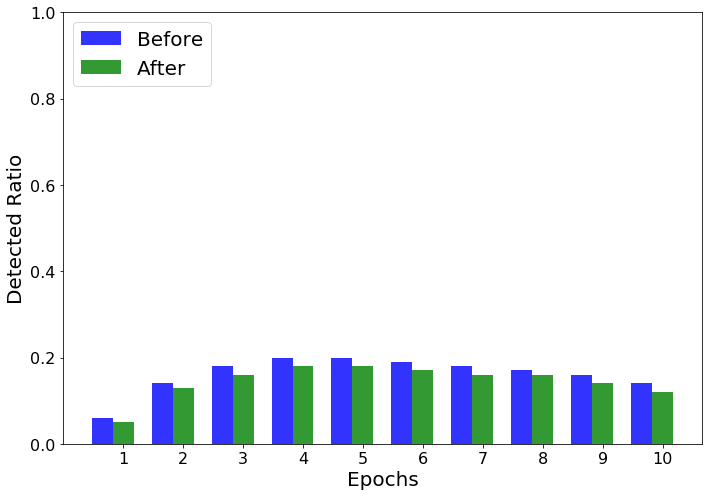}}
	\caption{Detection rates of objects before and after training on their corresponding images for (a) labeled instances and (b) instances with labels withheld during training.}
	\label{fig:forgetting_plot}	
\end{figure*}

Detection rates for labeled and unlabeled objects over time are shown in Figure \ref{fig:forgetting_plot}.
As expected, the model learns to detect a higher percentage of labeled instances over time, and objects are overall more likely to be detected immediately after the detector trains on them.
Despite not having an explicit learning signal, unlabeled objects are still learned throughout training, but at a lower rate than labeled ones.
In contrast with labeled objects, unlabeled object detections are discouraged with each PN gradient, leading to a dip in overall detection rates immediately after training.
Despite this, overall detection rates of unlabeled objects grows through the first 5 epochs of training, implying a repeated cycle of learning unlabeled objects from other intra-class examples, forgetting them when explicitly trained against them, and then learning them again.
Given the undesirability of this forced suppression of detected objects, we seek a method to remedy this behavior. 

\begin{wrapfigure}{r}{.4\linewidth}
    \vspace{-12mm}
	\centering
	\includegraphics[width=0.34\textwidth]{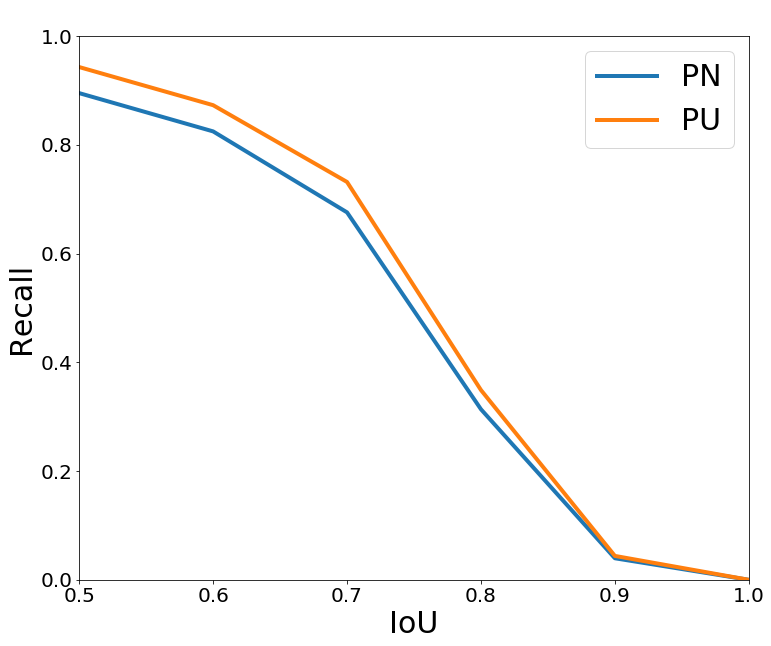}
	\caption{Recall of top 500 proposals from RPN after training on PASCAL VOC2007 when $\rho=0.5$.}
	\label{fig:recall}
	\vspace{-4mm}
\end{wrapfigure}

\section{Recall of Region Proposals}\label{apx:recall}
To investigate the effect of the PU risk estimator on the quality of the proposals from RPN stage, we examine the recall of the top 500 proposals, compared with the complete annotations. 
Higher recall means there are more proposals that match with the full-labeled annotations. 
In Figure \ref{fig:recall}, the recalls using the PU risk estimator are higher than those using the PN risk estimator. 
This illustrates the inclusion of more object proposals that are not included by the PN risk estimator because the corresponding ground truth annotations are missing. 

\section{OpenImages}\label{apx:openimages}
OpenImages \cite{Krasin2017,Kuznetsova2018} is a large object dataset consisting of 15.4 million bounding boxes from 600 classes across 1.9 million images.
In order to achieve its scale, the labeling effort was crowd-sourced to a large number of human annotators.
As pointed out in \cite{Everingham2010}, even increasing from 10 classes of objects in PASCAL VOC2006 to the 20 in VOC2007 resulted in a substantially larger number of labeling errors, as it became more difficult for human annotators to remember all of the object classes.
With 500 classes, this problem is worse by an order of magnitude for OpenImages.
While the creators of OpenImages designed an annotator training process to insure quality, there still are many examples of missing labels.
As such, PU learning as proposed is especially appropriate, even when considering full labels.

As in Section \ref{sec:PUvPN_VOC_COCO}, we train a ResNet101 Faster R-CNN object detector with both PN and PU classification losses.
Given the large size of the dataset, we restrict our analysis to 50 of the most prevalent classes, and subsample 140K images from the dataset containing at least one of the selected classes.
Of these 140K images, we train on 100K with \textit{full} annotations, and hold out 10K for validation and 30K as our test split.
We observe that, all other things equal, switching to our proposed PU approach results in an increase of $+3.0$, $+3.0$, and $+5.0$ for mAPs with IoU thresholds $\{0.25, 0.50, 0.75\}$ (see Table \ref{tab:OpenImages}).

\begin{table}[h]
    \caption{Detector performance on a subset of OpenImages at various IoU thresholds.}
	\begin{center}
	    \resizebox{0.4\columnwidth}{!}{%
		\begin{tabular}{l||c c c} 
			Method & $\text{AP}_{25}$ & $\text{AP}_{50}$ & $\text{AP}_{75}$\\
			\hline \hline
			PN & 37.7 & 33.6 & 20.7 \\
			PU & \textbf{40.7} & \textbf{36.6} & \textbf{25.7} \\
		\end{tabular}
		}
	\end{center}
	\label{tab:OpenImages}
\end{table}
\end{document}